\documentclass[conference]{IEEEtran}
\IEEEoverridecommandlockouts

\usepackage{cite}
\usepackage{textcomp}
\usepackage{amsmath,amssymb,amsfonts}
\usepackage{algorithm}
\usepackage{algorithmic}
\usepackage{graphicx}
\usepackage{xcolor}
\usepackage{subcaption}
\usepackage{booktabs}
\usepackage{multirow}
\usepackage{makecell}
\usepackage{array}
\usepackage{enumitem}

\def\BibTeX{{\rm B\kern-.05em{\sc i\kern-.025em b}\kern-.08em
    T\kern-.1667em\lower.7ex\hbox{E}\kern-.125emX}}

\begin{document}

\title{\mbox{Beyond the Pre-Service Horizon: Infusing In-Service}\\ 
Behavior for Improved Financial Risk Forecasting
}

\author{
\IEEEauthorblockN{Senhao Liu\textsuperscript{*}}
\IEEEauthorblockA{\textit{Renmin University of China} \\
Beijing, China \\
liusenhao@ruc.edu.cn}
\and
\IEEEauthorblockN{Zhiyu Guo\textsuperscript{*}}
\IEEEauthorblockA{\textit{Tencent Weixin Group} \\
Shenzhen, China \\
zyguo1999@gmail.com}
\and
\IEEEauthorblockN{Zhiyuan Ji}
\IEEEauthorblockA{\textit{Renmin University of China} \\
Beijing, China \\
zhiyuanji@ruc.edu.cn}
\and
\IEEEauthorblockN{Yueguo Chen\textsuperscript{\textdagger$\ddagger$}}
\IEEEauthorblockA{\textit{Renmin University of China} \\
Beijing, China \\
chenyueguo@ruc.edu.cn}
\and
\IEEEauthorblockN{Yateng Tang}
\IEEEauthorblockA{\textit{Tencent Weixin Group} \\
Shenzhen, China \\
fredyttang@tencent.com}
\and
\IEEEauthorblockN{Yunhai Wang}
\IEEEauthorblockA{\textit{Renmin University of China} \\
Beijing, China \\
wang.yh@ruc.edu.cn}
\and
\IEEEauthorblockN{Xuehao Zheng}
\IEEEauthorblockA{\textit{Tencent Weixin Group} \\
Shenzhen, China \\
xuehaozheng@tencent.com}
\and
\IEEEauthorblockN{Xiang Ao\textsuperscript{\textdagger\S}}
\IEEEauthorblockA{\textit{University of Chinese Academy of Sciences} \\
Beijing, China \\
aoxiang@ucas.ac.cn}
\thanks{* These authors contributed equally.}
\thanks{\textdagger~Corresponding authors.}
\thanks{$\ddagger$~Yueguo Chen is also with the Big Data and Responsible AI for National Governance, Renmin University of China, Beijing, China.}
\thanks{\S~Xiang Ao is also with the Institute of Intelligent Computing Technology, Chinese Academy of Sciences, Suzhou, China.}
}

\maketitle

\begin{abstract}

Typical financial risk management involves distinct phases for pre-service risk assessment and in-service default detection, often modeled separately. This paper proposes a novel framework, Multi-Granularity Knowledge Distillation (abbreviated as MGKD), aimed at improving pre-service risk prediction through the integration of in-service user behavior data. MGKD follows the idea of knowledge distillation, where the teacher model, trained on historical in-service data, guides the student model, which is trained on pre-service data. By using soft labels derived from in-service data, the teacher model helps the student model improve its risk prediction prior to service activation. Meanwhile, a multi-granularity distillation strategy is introduced, including coarse-grained, fine-grained, and self-distillation, to align the representations and predictions of the teacher and student models. This approach not only reinforces the representation of default cases but also enables the transfer of key behavioral patterns associated with defaulters from the teacher to the student model, thereby improving the overall performance of pre-service risk assessment. Moreover, we adopt a re-weighting strategy to mitigate the model's bias towards the minority class. Experimental results on large-scale real-world datasets from Tencent Mobile Payment demonstrate the effectiveness of our proposed approach in both offline and online scenarios.

\end{abstract}


\section{Introduction}\label{sec:intro}

Risk management has been central to financial services, permeating every stage of their lifecycle. This typically includes risk assessment before service provision and default detection during the service period, among other processes. Over the past few decades, machine learning models have been extensively employed in financial risk management, evolving from early expert systems~\cite{hodgkinson2003expert} and statistical machine learning~\cite{galindo2000credit} to current deep learning~\cite{mai2019deep,gan2019,liu2022user,guo2023transductive,qiao2025online} and graph-based machine learning approaches~\cite{temgnn2021,10.1145/3442381.3449989,10.1145/3534678.3539484,yuan2025dynamic,li2024boosting,liu2023flood,li2023revisiting,liu2022ud}.

\begin{figure}[!t]
    \centering
    \includegraphics[scale=0.203]{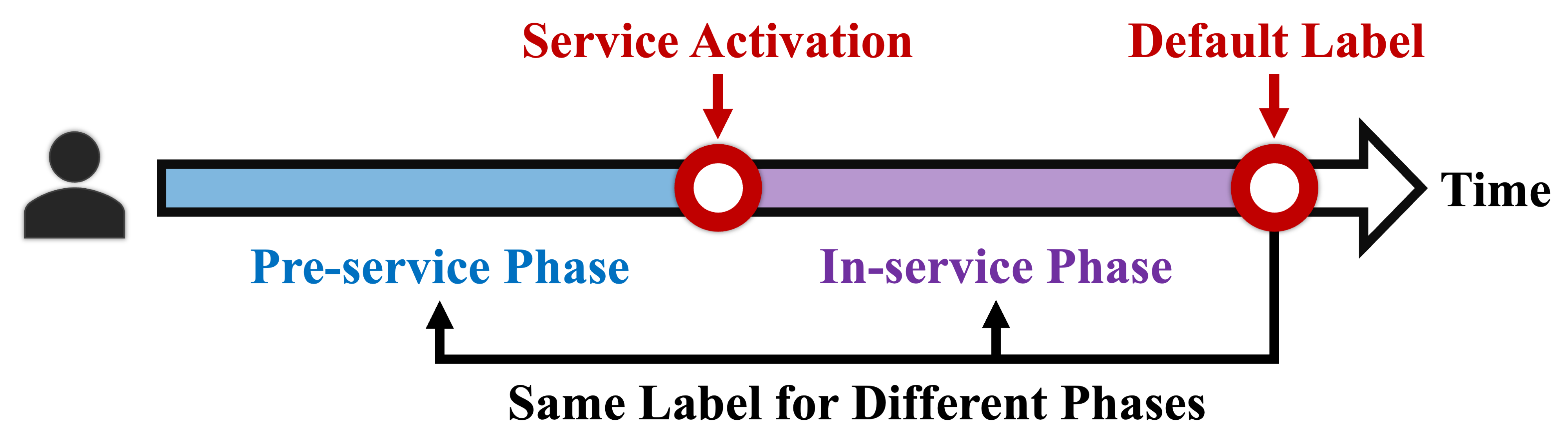}
    \caption{Pre-service and in-service phases in financial risk forecasting. In both phases, the same default label is employed for training the financial risk forecasting models.}
    \label{fig:1}
\end{figure}

Regardless of technological advancements, financial risk control based on machine learning techniques is generally divided into distinct phases aligned with the stages of service delivery. Figure~\ref{fig:1} exhibits an illustrative example of the phased approach in financial risk control. To assess whether a user qualifies for activation of a financial service, historical behavior data before service initiation is typically utilized to train models that predict the likelihood of future risk, such as the probability of default. Once the service is activated, a subsequent phase begins, during which the possibility of default is continuously evaluated using the user's behavior data collected during the service period.

\begin{figure}[!t]
    \centering
    \includegraphics[scale=0.61]{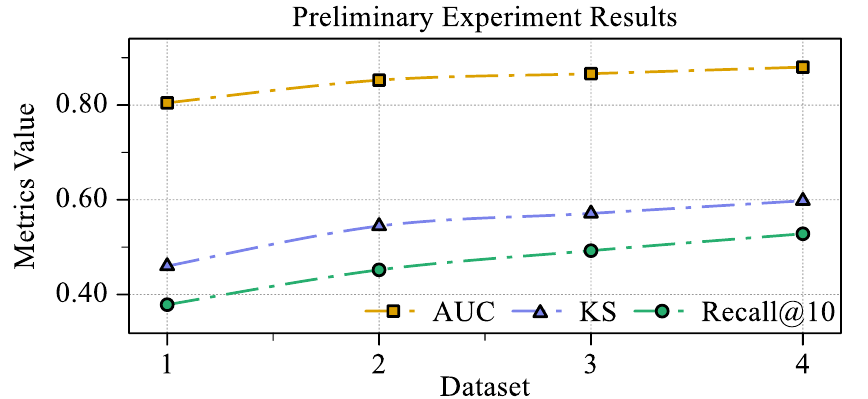}
    \caption{Preliminary experiment results. Model performance improves with the inclusion of in-service data for training. Dataset \textbf{1}: Pre-service only, \textbf{2}: Pre-service + In-service~(30 days), \textbf{3}: Pre-service + In-service~(60 days), \textbf{4}: Pre-service + In-service~(90 days).}
    \label{2}
\end{figure}

As shown in Figure~\ref{fig:1}, we can observe that both pre-service and in-service phases ultimately depend on the same outcome to train models, specifically, whether the user defaults after obtaining the service. 
As a result, an intuitive question arises in our study: Can we combine both pre-service and in-service data to train a stronger pre-service model? 
We verified this hypothesis through a preliminary experiment that incrementally incorporates in-service data into pre-service models to make predictions. 
Figure~\ref{2} shows a clear performance increase as more in-service data is incorporated into the model. This supports our idea that in-service data may provide useful information for identifying default users.

Based on these observations, we propose leveraging in-service data to develop a more accurate pre-service risk prediction model. However, several challenges must be addressed to achieve this goal. 
Firstly, in-service data cannot be captured and accessed prior to service activation. Hence, we need to prevent data leakage during model training and design strategies for transferring knowledge from the in-service data to the pre-service stage. 
Secondly, our objective is to identify the minority class of potential defaulters before activating financial services without raising the access threshold. 
Therefore, effectively extracting and transferring features related to minority defaulters from the in-service data becomes necessary.

To tackle these challenges, we propose a novel framework named \textit{Multi-Granularity Knowledge Distillation} (\textbf{MGKD}) for financial risk forecasting before the service activation. 
Specifically, to address the first challenge, the proposed MGKD follows the idea of teacher-student knowledge distillation. It infuses patterns of in-service user behavior into the pre-service prediction model by using historical user behaviors with both pre-service and in-service data. A teacher model is trained with the historical in-service data of a user, while a student model is trained with the pre-service data of the same user. The teacher model adopts soft labels derived from in-service data to guide the student model in calibrating its final prediction, thereby enhancing the risk prediction capability before service activation. 
To address the second challenge, we develop a multi-granularity distillation strategy. This approach uses coarse-grained distillation to align representations and fine-grained distillation to align predictions between the teacher and student models. Additionally, self-distillation is employed to further refine the model by utilizing its own predictions during training. 
These techniques not only strengthen the representations of default samples but also facilitate the transfer of indicative behavior patterns of defaulters between the teacher and student models, thereby further enhancing the overall performance of pre-service risk assessment.

The contributions of this work are summarized as follows:
\begin{itemize}
    \item We pioneer joint modeling of pre-service and in-service data for user-level financial risk forecasting, which is crucial for risk management in financial institutions.
    \item We propose a novel method named MGKD, employing multi-granularity knowledge distillation techniques to extract patterns related to defaulters' behaviors from in-service data to enhance pre-service risk prediction.
    \item We conducted experiments on a real-world dataset from Tencent Mobile Payment, revealing substantial improvements in AUC, KS, and Recall@10 metrics. The results validate the enhanced predictive performance and practical utility of our proposed approach.
\end{itemize}

\section{Related Work}\label{sec:related}


\subsection{Pre-service Financial Risk Prediction}

Financial risk forecasting usually decides whether a financial service activation application can be approved. Earlier research conducted by \cite{https://doi.org/10.1111/j.1540-6261.1968.tb00843.x} and \cite{Hand_Henley_1997} focused on linear statistical methods to predict bankruptcy and credit risk.  
Banks in traditional financial services have access to comprehensive customer data, including assets and income, which aids accurate credit risk assessment. In contrast, online financial platforms often struggle with limited user engagement, resulting in scarce informative features. This data deficiency impedes the precise modeling of user risk when relying solely on a single type of information.  
Subsequently, recent works have extended the application of deep neural networks to specific default detection tasks, such as fraud~\cite{Liu_Zhong_Ao_Sun_Lin_Feng_He_Tang_2020,Paasch_2014,Wang_Qi_Lin_Cui_Jia_Wang_Fang_Yu_Zhou_Yang_2019,2020Spatio,sahrnn}, cash-out~\cite{hu2019cash}, and malicious behavior~\cite{Liu_Chen_Yang_Zhou_Li_Song_2018, Wang_Wen_Wu_Huang_Xiong_2019}, showing that relational data enhances default identification~\cite{Zhong_Liu_Ao_Hu_Feng_Tang_He_2020}.  
Due to the richer information modeled in graph data, recent research has also begun to focus on GNN-based methods~\cite{ding2025spear,huang2022auc,guo2025grasp}. For example, \cite{cheng2023anti} proposed a group-aware deep graph learning approach for anti-money laundering. \cite{ICDM22devil} introduced a Disentangled Information Graph Neural Network~(DIGNN) to tackle the inconsistency problem in graph-based fraud detection.

\subsection{In-service Default Account Detection}

In-service default account detection predicts whether a user's behavior will be deemed fraudulent after service activation. Traditional methods extract statistical features from user profiles and historical behaviors, using supervised learning approaches like logistic regression, SVM, and neural networks. Recent research leverages graph-based methods to model complex user interactions in financial contexts, improving the accuracy and robustness of default detection. 
For example, \cite{Liu_Chen_Yang_Zhou_Li_Song_2018} introduced a GNN to detect malicious accounts by learning discriminative embeddings from heterogeneous account-device graphs. \cite{hu2019cash} proposed a meta-path-based graph embedding method for user cash-out prediction, using meta-path-based neighbors to aggregate user attributes and structural information. 
\cite{10.1145/3442381.3449989} proposed PC-GNN, which addresses class imbalance in graph-based fraud detection by using label-balanced sampling and neighborhood selection. \cite{10.1145/3340531.3412724} constructed a multiplex graph to detect loan defaults by capturing multifaceted relationships among users. \cite{10.1145/3580305.3599351} proposed a motif-preserving GNN combined with curriculum learning to exploit both lower- and higher-order graph structures for financial default detection. 
Some works have also explored data synthesis techniques to address challenges such as class imbalance.
However, these works rely solely on single-phase information and overlook the connection between pre-service and in-service stages, distinguishing our approach from existing methods.





\begin{figure*}[!t]
  \centering
  \includegraphics[width=\textwidth]{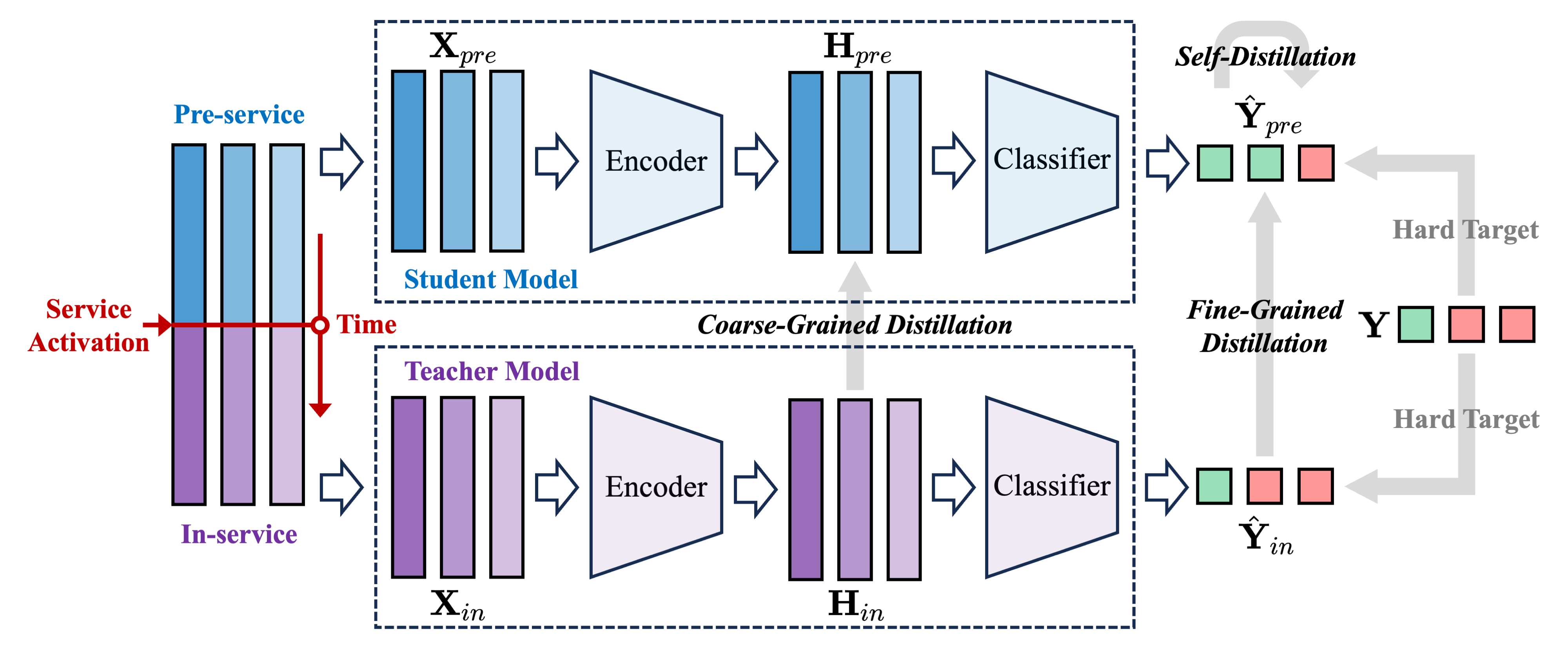} 
  \caption{Illustration of our proposed MGKD framework: The teacher model $f_{in}$ is trained using in-service features $\mathbf{X}_{in}$ and default labels $\mathbf{Y}$, while the student model $f_{pre}$ is trained using pre-service features $\mathbf{X}_{pre}$ and default labels $\mathbf{Y}$. The student model $f_{pre}$ is enhanced by injecting knowledge from the teacher model $f_{in}$ through coarse-grained, fine-grained, and self-distillation. Coarse-grained distillation aligns user representations $\mathbf{H}_{in}$ and $\mathbf{H}_{pre}$, while fine-grained distillation aligns default risk predictions $\hat{\mathbf{Y}}_{in}$ and $\hat{\mathbf{Y}}_{pre}$. In addition, self-distillation aligns the student model's current default risk predictions $\hat{\mathbf{Y}}_{\mathit{pre}}$ with its earlier predictions $\hat{\mathbf{Y}}_{\mathit{pre}}'$.}
  \label{Figure-Model}
\end{figure*}

\section{Problem Statement} \label{sec:problem}

Assume that the pre-service data has been processed into vectorized features $\mathbf{X}_{pre} \in \mathbb{R}^{N \times D}$, where $N$ is the number of users and $D$ is the feature dimension. Financial service platforms monitor users who have activated the service, recording whether they default during the service period and generating labels $\mathbf{Y} \in \{0, 1\}^{N}$, where $\mathbf{Y}_i = 1$ indicates default and $\mathbf{Y}_i = 0$ otherwise. Using $\mathbf{X}_{pre}$ and $\mathbf{Y}$, a default risk assessment model $f_{pre}$ can be trained to score new applicants. Unlike existing methods, we also utilize the in-service data after service activation, which has been processed into vectorized features $\mathbf{X}_{in} \in \mathbb{R}^{N \times D}$, to develop an enhanced default risk assessment model $f_{pre}^{\prime}$, i.e., $(\mathbf{X}_{pre}, \mathbf{X}_{in}, \mathbf{Y}) \to f_{pre}^{\prime}$. During testing, $f_{pre}^{\prime}$ will only require $\mathbf{X}_{pre}$ as input, enabling its application in assessing the default risk of new applicants before service activation.

\section{Methodology}\label{sec:model}

In this section, we elaborate on our MGKD framework.

\subsection{Overall Framework}

As illustrated in Figure~\ref{Figure-Model}, our MGKD framework first trains the teacher model $f_{\mathit{in}}$ using the in-service data.
Then, we train the student model $f_{\mathit{pre}}$ using the pre-service data, simultaneously injecting knowledge from the in-service model $f_{\mathit{in}}$ through a multi-granularity knowledge distillation strategy.
We ultimately obtain an enhanced pre-service model $f_{\mathit{pre}}^{\prime}$ that only requires the pre-service data (vectorized features $\mathbf{X}_{pre}$) as input, making it suitable for assessing the default risk of new applicants before service activation.

\subsection{Training Teacher Model}

Traditional knowledge distillation frameworks involve a trained teacher model and an untrained student model. Both models receive the same input, with the teacher's outputs supervising the student's training.
Unlike traditional knowledge distillation paradigms, our approach utilizes behavioral features from distinct stages in separate models. Specifically, the teacher model processes in-service data, while the student model processes pre-service data. This methodology is driven by our objective to assess user default risk using only pre-service features. In contrast, in-service features offer richer information and more precise risk characterization.
Consequently, we aim to integrate knowledge from in-service features into the pre-service default risk assessment model to enhance its performance.

In more detail, we first train an in-service default risk assessment model $f_{\mathit{in}}$ as the teacher model using in-service features $\mathbf{X}_{\mathit{in}}$ and default labels $\mathbf{Y}$. $f_{\mathit{in}}$ takes $\mathbf{X}_{\mathit{in}}$ as input and outputs the predicted default probabilities $\hat{\mathbf{Y}}_{\mathit{in}} \in [0, 1]^{N}$. The model is trained using the Kullback-Leibler divergence as the loss function:
\begin{equation}
\mathcal{L}_{\mathit{in}} = \mathrm{KL}(\mathbf{Y} \parallel \hat{\mathbf{Y}}_{\mathit{in}}).
\end{equation}
After training, we obtain a trained in-service default risk assessment model $f_{\mathit{in}}$. This model will serve as the teacher model to guide the training of the pre-service default risk assessment model $f_{\mathit{pre}}$, which serves as the student model.

\subsection{Coarse-Grained Distillation}

In our approach, we first focus on aligning the representations generated by the teacher and student models at a coarse-grained level to facilitate knowledge transfer. To achieve this, the in-service model $f_{\mathit{in}}$ is decomposed into an encoder $g_{\mathit{in}}$ and a classifier $h_{\mathit{in}}$, and the student pre-service model $f_{\mathit{pre}}$ is decomposed into an encoder $g_{\mathit{pre}}$ and a classifier $h_{\mathit{pre}}$, satisfying the following expressions:
\begin{equation}
f_{\mathit{in}} = h_{\mathit{in}} \circ g_{\mathit{in}}, \qquad f_{\mathit{pre}} = h_{\mathit{pre}} \circ g_{\mathit{pre}}.
\end{equation}
By decomposing the model, we can obtain not only the model's predictions of user default risk but also the representations generated for users. Specifically, the representations from the in-service model and the pre-service model are denoted as $\mathbf{H}_{\mathit{in}} \in \mathbb{R}^{N \times d}$ and $\mathbf{H}_{\mathit{pre}} \in \mathbb{R}^{N \times d}$, respectively, satisfying $\mathbf{H}_{\mathit{in}} = g_{\mathit{in}}(\mathbf{X}_{\mathit{in}})$ and $\mathbf{H}_{\mathit{pre}} = g_{\mathit{pre}}(\mathbf{X}_{\mathit{pre}})$, where $d$ denotes the dimensionality of these representations.

Recall that we aim to align the user representations generated by the pre-service model with those from the in-service model, thus injecting the knowledge from in-service features into the pre-service model. To measure the alignment between the in-service and pre-service representations, we introduce the representation alignment loss function $\mathcal{L}_{\mathit{feat}}$:
\begin{equation}
\mathcal{L}_{\mathit{feat}} = D(\mathbf{H}_{\mathit{in}}, \mathbf{H}_{\mathit{pre}}),
\end{equation}
where $D(\cdot, \cdot)$ can be mean squared error, cosine distance, etc. $\mathbf{H}_{\mathit{in}}$ is generated by the trained in-service model $f_{\mathit{in}}$, which is used to guide the pre-service model in generating the user representations $\mathbf{H}_{\mathit{pre}}$.

\subsection{Fine-Grained Distillation}

In the classic knowledge distillation framework, the training process of the student model is supervised by the predictions of the teacher model, thereby transferring knowledge from the teacher to the student model. In our context, we obtain soft labels $\hat{\mathbf{Y}}_{\mathit{in}} \in [0, 1]^{N}$ for each user's default risk from the trained in-service model $f_{\mathit{in}}$. To utilize both hard labels $\mathbf{Y}$ and soft labels $\hat{\mathbf{Y}}_{\mathit{in}}$ for training the pre-service model, we introduce the hard target loss function $\mathcal{L}_{\mathit{hard}}$ and the soft target loss function $\mathcal{L}_{\mathit{soft}}$, which are then combined into the target loss function $\mathcal{L}_{\mathit{label}}$:
\begin{equation}
\mathcal{L}_{\mathit{hard}} = \mathrm{KL}(\mathbf{Y} \parallel \hat{\mathbf{Y}}_{\mathit{pre}}),
\end{equation}
\begin{equation}
\mathcal{L}_{\mathit{soft}} = \mathrm{KL}\left({\hat{\mathbf{Y}}_{\mathit{in}}} / {\tau} \parallel {\hat{\mathbf{Y}}_{\mathit{pre}}} / {\tau}\right),
\end{equation}
\begin{equation}
\mathcal{L}_{\mathit{label}} = (1 - \alpha) \mathcal{L}_{\mathit{hard}} + \alpha \mathcal{L}_{\mathit{soft}}.
\end{equation}
In the above expressions, the hard target loss function $\mathcal{L}_{\mathit{hard}}$ uses KL divergence to align the predicted labels $\hat{\mathbf{Y}}_{\mathit{pre}}$ with the hard labels $\mathbf{Y}$. The soft target loss function $\mathcal{L}_{\mathit{soft}}$ employs KL divergence with a temperature coefficient $\tau \ge 1$ to align the predicted labels $\hat{\mathbf{Y}}_{\mathit{pre}}$ with the soft labels $\hat{\mathbf{Y}}_{\mathit{in}}$. The temperature coefficient is used to smooth the labels, helping the student model gain more information about the classes and enhance knowledge transfer. The target loss function $\mathcal{L}_{\mathit{label}}$ is a weighted combination of these two loss functions, where $\alpha \in [0, 1]$ is the weighting factor. We refer to this process as fine-grained knowledge distillation.

\subsection{Self-Distillation}

To further enhance the performance of the student model, we introduce a self-distillation strategy. This strategy enables the student model to refine its own predictions by leveraging its output probabilities during training. The self-distillation loss is denoted as $\mathcal{L}_{\mathit{self}}$:
\begin{equation}
\mathcal{L}_{\mathit{self}} = \mathrm{KL}\!\left({\hat{\mathbf{Y}}_{\mathit{pre}}} / {\tau} \parallel {\hat{\mathbf{Y}}_{\mathit{pre}}^{\prime}} / {\tau}\right),
\end{equation}
where $\hat{\mathbf{Y}}_{\mathit{pre}}$ represents the softened predictions of the student model, $\hat{\mathbf{Y}}_{\mathit{pre}}^{\prime}$ denotes the corresponding outputs from a previous training epoch. The self-distillation loss $\mathcal{L}_{\mathit{self}}$ is combined with the existing loss functions to jointly optimize the student model, contributing to more accurate and robust predictions.

\subsection{Model Training}

To summarize the entire process of our proposed MGKD method, we first train an in-service default risk prediction model $f_{\mathit{in}}$ as the teacher model using in-service features $\mathbf{X}_{\mathit{in}}$. 
Next, we train a pre-service default risk prediction model $f_{\mathit{pre}}$ as the student model using pre-service features $\mathbf{X}_{\mathit{pre}}$. 
During the training of $f_{\mathit{pre}}$, we aim to inject knowledge from $f_{\mathit{in}}$ to enhance its performance. 
Therefore, we introduce a multi-granularity knowledge distillation strategy. 
This strategy aligns the user representations $\mathbf{H}_{\mathit{in}}$ and $\mathbf{H}_{\mathit{pre}}$ of the two models through coarse-grained distillation, while simultaneously aligning the user default risk predictions $\hat{\mathbf{Y}}_{\mathit{in}}$ and $\hat{\mathbf{Y}}_{\mathit{pre}}$ through fine-grained distillation. In the process of aligning predictions, we combine hard target loss and soft target loss to balance classification accuracy with learning from the teacher model. In addition, we incorporate self-distillation to further refine the predictions by leveraging outputs from earlier training epochs. Finally, we weight and combine the knowledge distillation losses at different granularities to obtain the overall knowledge distillation loss:
\begin{equation}
\mathcal{L}_{\mathit{distill}} = \mathcal{L}_{\mathit{label}} + \beta \mathcal{L}_{\mathit{feat}} + \lambda \mathcal{L}_{\mathit{self}},
\end{equation}
where $\beta \ge 0$ and $\lambda \ge 0$ are weighting factors used to balance the different granularities in knowledge distillation. 
By using the above framework to train the student model, we ultimately obtain a pre-service default risk assessment model $f_{\mathit{pre}}^{\prime}$, which is enhanced with in-service behavioral data $\mathbf{X}_{\mathit{in}}$. 
The enhanced model $f_{\mathit{pre}}^{\prime}$ only takes pre-service behavioral data $\mathbf{X}_{\mathit{pre}}$ as input, making it suitable for assessing the default risk of new financial service applicants before service activation.

Additionally, we adopt a re-weighting strategy and focal loss to mitigate the model's bias towards the minority class. Details are reported in the Appendix.

\section{Experiment Setup}\label{sec:setup}

\subsection{Dataset}
We collect two real-world datasets\footnote{The dataset in this paper is properly sampled only for testing purposes and does not imply any commercial information. All users' private information is removed from the dataset. Moreover, the experiment was conducted locally on Tencent's server by formal employees who strictly followed data protection regulations.\label{claim}} (denoted as \textbf{S1} and \textbf{S2}) from Tencent Mobile Payment on the premise of complying with the security and privacy policies.

\begin{itemize}
    \item The first dataset, \textbf{S1}, is used for the main experiments (\textbf{RQ1}, \textbf{RQ3}, and \textbf{RQ4}). It includes 387,732 users for training (2020/03/24 to 2021/05/19), 36,021 users for validation (2021/05/20 to 2021/05/31), and 35,391 users for testing (2021/06/01 to 2021/06/16). Positive samples are fraudulent users, while negative samples are benign, with a positive rate of approximately 10.0\%.
    \item The second dataset, \textbf{S2}, contains approximately 900,000 samples and is chronologically later than \textbf{S1}. It is used for the online simulation experiment (\textbf{RQ2}) to evaluate the MGKD framework's robustness and generalizability under real-world conditions with significant data shifts. Detailed statistics for \textbf{S1} and \textbf{S2} are shown in Table \ref{dataset}.
\end{itemize}

\subsection{Compared Methods}
To evaluate the effectiveness of our proposed MGKD framework, we compare it with several baseline models, including traditional machine learning methods, ensemble techniques, graph-based deep learning methods, and different variants of our framework.

\begin{enumerate}
    \item \textbf{Traditional Machine Learning Methods}:
    \begin{itemize}
        \item \textbf{LR (Logistic Regression)}~\cite{kleinbaum2002logistic}: A simple linear model for binary classification.
        \item \textbf{SVM (Support Vector Machine)}~\cite{hearst1998support}: Finds a hyperplane for both linear and non-linear data separation.
    \end{itemize}
    
    \item \textbf{Gradient Boosting Methods}:
    \begin{itemize}
        \item \textbf{XGBoost}~\cite{chen2016xgboost}: A decision tree-based method balancing accuracy and efficiency.
        \item \textbf{LightGBM}~\cite{ke2017lightgbm}: Optimized for speed and memory usage using histogram techniques.
        \item \textbf{CatBoost}~\cite{prokhorenkova2018catboost}: Handles categorical features well, reducing preprocessing needs.
    \end{itemize}
    
    \item \textbf{Graph-based Deep Learning Methods}:
    \begin{itemize}
        \item \textbf{GCN (Graph Convolutional Network)}~\cite{kipf2016semi}: Captures neighborhood information through convolution.
        \item \textbf{GAT (Graph Attention Network)}~\cite{velivckovic2017graph}: Uses attention mechanisms to weigh neighboring nodes.
        \item \textbf{GraphSAGE}~\cite{hamilton2017inductive}: Learns node embeddings by aggregating information from sampled neighbors.
    \end{itemize}

    \item \textbf{Variants of Our Proposed Framework}:
    \begin{itemize}
        \item \textbf{Oracle}: An MLP model using both pre-service and in-service data for an \textbf{upper bound} reference.
        \item \textbf{Pre-service Model}: An MLP baseline relying solely on pre-service data, acting as a \textbf{lower bound}.
        \item \textbf{MGKD Framework}: Full implementation of our Multi-Granularity Knowledge Distillation (MGKD) framework, which leverages both pre-service and in-service data for improved risk forecasting.
    \end{itemize}
\end{enumerate}

\begin{table}[!t]
\centering
\caption{Statistical information of datasets S1 and S2.}
\renewcommand{\arraystretch}{1.5} 
\setlength{\tabcolsep}{1mm} 
\begin{tabular}{@{}>{\centering\arraybackslash}m{0.8cm}|>{\centering\arraybackslash}m{1.0cm}|>{\centering\arraybackslash}m{1.3cm}|>{\centering\arraybackslash}m{1.5cm}|>{\centering\arraybackslash}m{1.4cm}|>{\centering\arraybackslash}m{1.3cm}@{}}
\toprule
\textbf{Data} & \textbf{Subset} & \textbf{Positive} & \textbf{Negative} & \textbf{Total} & \textbf{Positive Rate} \\
\midrule
\multirow{4}{*}{\textbf{S1}} 
& Train   & 27,836 & 359,896 & 387,732 & 7.2\% \\
& Valid & 4,378  & 31,643  & 36,021  & 12.2\% \\
& Test    & 4,330  & 31,061  & 35,391  & 12.2\% \\
& \textbf{Total} & \textbf{36,544} & \textbf{422,600} & \textbf{459,144} & \textbf{7.97\%} \\
\midrule
\multirow{4}{*}{\textbf{S2}} 
& Train   & 76,453 & 839,571 & 916,024 & 8.35\% \\
& Valid & 4,558  & 41,011  & 45,569  & 10.00\% \\
& Test    & 5,364  & 43,591  & 48,955  & 10.96\% \\
& \textbf{Total} & \textbf{86,375} & \textbf{924,173} & \textbf{1,010,548} & \textbf{8.55\%} \\
\bottomrule
\end{tabular}
\label{dataset}
\end{table}

\subsection{Metrics and Reproducibility}
We evaluate the compared models using AUC, KS, and Recall@$k$~($k=10$ in experiments). Higher values indicate better performance. The proposed model is implemented in PyTorch~\cite{paszke2017automatic}, and experiments are conducted on an NVIDIA A100 GPU server. The model includes two hidden layers, each with 256 dimensions. Knowledge distillation uses parameters $\alpha = 0.2$, $\beta = 0.25$, and a temperature coefficient $\tau = 2.5$. Training is performed with a batch size of $100,000$ using the Adam optimizer (learning rate $0.0005$), a weight decay of $10^{-6}$, and a dropout rate of $0.4$. The model is trained for up to 100 epochs with early stopping triggered after 50 epochs without validation loss improvement. For graph-based methods, we construct a KNN graph using cosine similarity with top-$k=20$.

\begin{table*}[!t]
\caption{Offline Experiment Results. The proposed MGKD method significantly outperforms a variety of advanced baselines that do not leverage in-service data, across all evaluation metrics. Moreover, it effectively narrows the performance gap with the theoretical upper bound (Oracle) that explicitly uses both pre-service and in-service data.}
\begin{center}
\renewcommand{\arraystretch}{1.5} 
\setlength{\tabcolsep}{3pt} 
\begin{tabular}{m{3.5cm}<{\centering} | m{1.1cm}<{\centering} m{1.2cm}<{\centering} m{1.8cm}<{\centering} | m{1.1cm}<{\centering} m{1.2cm}<{\centering} m{1.8cm}<{\centering} | m{1.1cm}<{\centering} m{1.2cm}<{\centering} m{1.8cm}<{\centering}}
\cmidrule(lr){1-10}
\multirow{2}{*}{\raisebox{-1.0ex}{Method}} & \multicolumn{3}{c|}{30 days} & \multicolumn{3}{c|}{60 days} & \multicolumn{3}{c}{90 days} \\
\cmidrule(lr){2-10}
& AUC & KS & Recall@10 & AUC & KS & Recall@10 & AUC & KS & Recall@10 \\
\cmidrule(lr){1-10}
Oracle & 0.8526 & 0.5446 & 0.4520 & 0.8663 & 0.5712 & 0.4925 & 0.8801 & 0.5979 & 0.5282 \\
\cmidrule(lr){1-10}
LR & 0.7814 & 0.4216 & 0.3409 & 0.7814 & 0.4216 & 0.3409 & 0.7814 & 0.4216 & 0.3409 \\
SVM & 0.7918 & 0.4421 & 0.3578 & 0.7918 & 0.4421 & 0.3578 & 0.7918 & 0.4421 & 0.3578 \\
XGBoost & 0.8057 & 0.4584 & 0.3834 & 0.8057 & 0.4584 & 0.3834 & 0.8057 & 0.4584 & 0.3834 \\
LightGBM & 0.8026 & 0.4638 & 0.3771 & 0.8026 & 0.4638 & 0.3771 & 0.8026 & 0.4638 & 0.3771 \\
CatBoost & 0.8055 & 0.4584 & 0.3831 & 0.8055 & 0.4584 & 0.3831 & 0.8055 & 0.4584 & 0.3831 \\
GCN & 0.7901 & 0.4288 & 0.3441 & 0.7901 & 0.4288 & 0.3441 & 0.7901 & 0.4288 & 0.3441 \\
GAT & 0.7839 & 0.4243 & 0.3413 & 0.7839 & 0.4243 & 0.3413 & 0.7839 & 0.4243 & 0.3413 \\
GraphSAGE & 0.7945 & 0.4432 & 0.3647 & 0.7945 & 0.4432 & 0.3647 & 0.7945 & 0.4432 & 0.3647 \\
\cmidrule(lr){1-10}
Pre-service & 0.8047 & 0.4600 & 0.3785 & 0.8047 & 0.4600 & 0.3785 & 0.8047 & 0.4600 & 0.3785 \\
MGKD & \textbf{0.8210} & \textbf{0.4885} & \textbf{0.4012} & \textbf{0.8193} & \textbf{0.4831} & \textbf{0.3992} & \textbf{0.8178} & \textbf{0.4751} & \textbf{0.3979} \\
\cmidrule(lr){1-10}
\# Difference w.r.t. Oracle & 0.0316 & 0.0561 & 0.0508 & 0.0470 & 0.0881 & 0.0933 & 0.0623 & 0.1228 & 0.1303 \\
\# Relative Improvement & \textbf{2.03\%} & \textbf{6.20\%} & \textbf{6.00\%} & \textbf{1.81\%} & \textbf{5.02\%} & \textbf{5.47\%} & \textbf{1.63\%} & \textbf{3.28\%} & \textbf{5.13\%} \\
\cmidrule(lr){1-10}
\end{tabular}
\label{offline}
\end{center}
\end{table*}

\section{Experiment Results}\label{sec:exp}

In this section, we display and analyze the results of our experiments on the real-world datasets to demonstrate the effectiveness of our proposed model. Particularly, we aim to address the following research questions:

\begin{itemize}
    \item \textbf{RQ1}: Does our proposed method improve the predictive performance compared to baseline models?
    \item \textbf{RQ2}: Can our proposed framework maintain performance robustness when deployed in an online setting?
    \item \textbf{RQ3}: How does each component of our approach contribute to the overall performance?
    \item \textbf{RQ4}: How sensitive is the MGKD framework's performance to changes in its key hyperparameters?
\end{itemize}

\subsection{Main Experiment~(RQ1)}
To answer the first research question, we compare the performance of our MGKD framework against different types of baseline methods on three subsets of \textbf{S1} with different time window lengths (30 days, 60 days, 90 days) to evaluate its effectiveness in default prediction. Table \ref{offline} presents the main results of our proposed approach compared to the baselines. Key findings are summarized below.

\subsubsection{MGKD vs. Traditional Methods}

We compare our MGKD framework with traditional non-deep learning methods, including LR, SVM, XGBoost, LightGBM, and CatBoost, as well as graph-based deep learning methods such as GCN, GAT, and GraphSAGE. As shown in Table \ref{offline}, MGKD outperforms all baseline models across all metrics and time windows. In the 30-day window, MGKD achieves an AUC of 0.8210, surpassing XGBoost, the best traditional method, by 1.90\%, and GraphSAGE, the best graph-based method, by 3.34\%. For KS, MGKD improves by 10.50\% over SVM, 15.87\% over LR, and 13.92\% over GCN. In Recall@10, MGKD exceeds CatBoost by 4.72\% and GraphSAGE by 10.00\%. These results demonstrate that traditional and graph-based methods, while effective in certain aspects, struggle to capture nuanced behaviors in in-service data. By integrating pre-service and in-service data through multi-granularity knowledge distillation, MGKD achieves superior performance.

\subsubsection{MGKD vs. Bound Model (Oracle and Pre-service Model)}
Next, we compare our framework against the boundary models, specifically the Oracle and Pre-service models, which represent the upper and lower bounds of performance, respectively.

\begin{itemize} 
    \item \textbf{Comparison with Oracle}: As expected, the Oracle model achieves the highest performance across all metrics since it has access to both historical and current data simultaneously. However, the MGKD framework reduces the performance disparity. For instance, in the 30-day window, the AUC of MGKD is 0.8210, which is only 3.72\% lower than the Oracle's AUC of 0.8526. The relative improvement in KS (6.20\%) and Recall@10 (6.00\%) over the Pre-service Model suggests that MGKD effectively leverages in-service data to close the gap with the Oracle.
    \item \textbf{Comparison with Pre-service Model}: The MGKD framework outperforms the Pre-service Model across all metrics, which solely relies on pre-service data. For the 30-day window, MGKD achieves a 2.03\% higher AUC, a 6.20\% improvement in the KS metric, and a 6.00\% increase in Recall@10. These results indicate that incorporating in-service data through the MGKD framework substantially improves the identification of high-risk cases.
\end{itemize}

The results demonstrate that while MGKD does not reach the theoretical limits of the Oracle, it significantly outperforms the Pre-service baseline. This enhanced performance is primarily attributed to the utilization of post-activation in-service features, providing more comprehensive insights into users' behavior over time. These findings suggest that MGKD is a practical and effective solution for real-world financial risk prediction tasks where real-time data is available.

Although the MGKD framework demonstrates improvements over the Pre-service Model across all three time windows, indicating its overall effectiveness, we still observe a slight decline in performance as the time window lengthens from 30 days to 90 days. This decline can be attributed to the temporal shift in user behavior. As the interval between behaviors increases, the relevance of the knowledge used to predict financial risk diminishes, and integrating this shifted knowledge into the model becomes less effective. Consequently, the model's ability to maintain high predictive accuracy is compromised with longer time windows.

\subsection{Online Simulation~(RQ2)}
Next, to evaluate the robustness of the MGKD framework in the presence of data shifts, we conducted an online simulation experiment on dataset \textbf{S2} from Tencent Mobile Payment. The dataset \textbf{S2}, consisting of approximately 900,000 new samples, is chronologically later than dataset \textbf{S1}. Given that the MGKD$_{30}$ configuration demonstrated the best performance in the preceding offline experiments, we employed only this specific model for the online simulation. This selection ensures that the most effective variant of our framework is evaluated under real-world conditions, providing a focused assessment of its robustness and applicability in an operational environment.

Table \ref{online2} highlights that incorporating in-service data from the past 30 days enhances the model's accuracy during the online testing period, similar to the results observed in offline experiments. In the first testing period, the MGKD framework improved AUC by 1.32\% and KS by 3.81\%. Similarly, in the second testing period, the framework maintained these improvements with AUC increasing by 1.11\% and KS by 3.76\%. These improvements demonstrate that the MGKD framework's effectiveness observed in offline experiments is sustained during online simulation, confirming the robustness of the model's enhancements.

\begin{table}[!t]
\centering
\caption{Online Simulation Experiment Results.}
\setlength{\tabcolsep}{2pt} 
\renewcommand{\arraystretch}{2} 
\begin{tabular}{@{} >{\centering\arraybackslash}m{2cm} >{\centering\arraybackslash}m{2.2cm} >{\centering\arraybackslash}m{1cm} >{\centering\arraybackslash}m{1cm} >{\centering\arraybackslash}m{1.6cm} @{}}
\toprule
Test Data Window & Method & AUC & KS & Recall@10 \\
\midrule
\multirow{3}{*}{\makecell{20211001 \\ - \\ 20211031}} & \raisebox{0.5ex}{\makecell{MGKD$_{30}$}} & \raisebox{0.5ex}{\textbf{0.7884}} & \raisebox{0.5ex}{\textbf{0.4302}} & \raisebox{0.5ex}{\textbf{0.3677}} \\
\cline{2-5}
& \raisebox{0.2ex}{Pre-service} & \raisebox{0.2ex}{0.7781} & \raisebox{0.2ex}{0.4144} & \raisebox{0.2ex}{0.3544} \\
\cline{2-5}
& \raisebox{-0.3ex}{\makecell{\# Relative \\ Improvement}} & \raisebox{-0.3ex}{\textbf{1.32\%}} & \raisebox{-0.3ex}{\textbf{3.81\%}} & \raisebox{-0.3ex}{\textbf{3.75\%}} \\
\midrule
\multirow{3}{*}{\makecell{20211101 \\ - \\ 20211206}} & \raisebox{0.5ex}{\makecell{MGKD$_{30}$}} & \raisebox{0.5ex}{\textbf{0.7853}} & \raisebox{0.5ex}{\textbf{0.4279}} & \raisebox{0.5ex}{\textbf{0.3567}} \\
\cline{2-5}
& \raisebox{0.2ex}{Pre-service} & \raisebox{0.2ex}{0.7767} & \raisebox{0.2ex}{0.4124} & \raisebox{0.2ex}{0.3476} \\
\cline{2-5}
& \raisebox{-0.3ex}{\makecell{\# Relative \\ Improvement}} & \raisebox{-0.3ex}{\textbf{1.11\%}} & \raisebox{-0.3ex}{\textbf{3.76\%}} & \raisebox{-0.3ex}{\textbf{2.62\%}} \\
\bottomrule
\end{tabular}
\label{online2}
\end{table}

Notably, the AUC exhibited the smallest decline when transitioning from offline to online experiments, with a decrease in improvement from 2.03\% (offline) to 1.11\%--1.32\% (online). This modest reduction suggests that AUC, which reflects the model's overall ranking quality, remains relatively stable across different testing environments. In contrast, both KS and Recall@10 experienced more significant declines, with KS dropping from 6.20\% (offline) to 3.76\%--3.81\% (online) and Recall@10 decreasing from 6.00\% (offline) to 2.62\%--3.75\% (online). These declines indicate that KS and Recall@10 are more sensitive to shifts in user behavior and data distribution, particularly in identifying high-risk individuals. These results underscore the importance of adapting predictive models to real-world variability. By leveraging in-service data, the MGKD framework enhances pre-service risk assessment, offering a more nuanced understanding of user behavior crucial for making informed and accurate financial decisions in online settings.

\begin{figure*}[t]
    \centering
    \includegraphics[width=\textwidth]{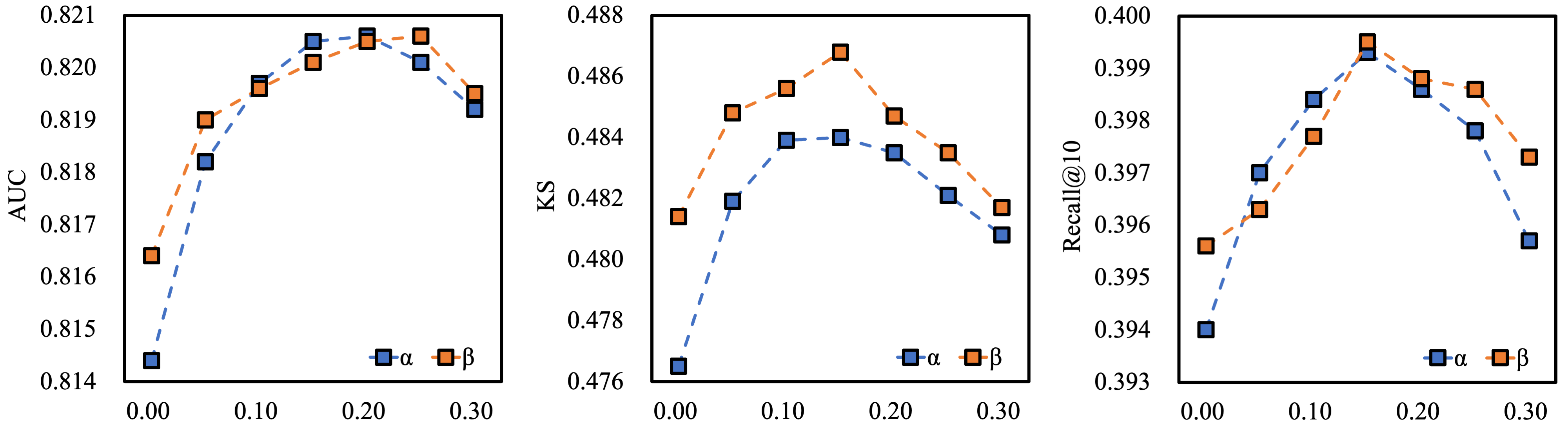}
    \caption{Hyperparameter Sensitivity Analysis. As the values of $\alpha$ or $\beta$ increase, all evaluation metrics (AUC, KS, and Recall@10) exhibit a trend of rising first and then falling, indicating a trade-off effect.}
    \label{sensitivity}
\end{figure*}

\subsection{Ablation Study~(RQ3)}
To address the third research question, we conducted an ablation study using the 30-day subset of data to evaluate the contribution of each component within the MGKD framework. Specifically, we assessed the impact of pre-training, coarse-grained distillation, and fine-grained distillation by removing each component in separate experiments. Table \ref{ablation_mgkd} shows that removing any component results in a decline in all performance metrics, highlighting the importance of each part in enhancing the overall predictive performance.

\begin{table}[t]
\caption{Ablation Study of MGKD Framework.}
\centering
\renewcommand{\arraystretch}{1.5}
\begin{tabular}{m{3cm}<{\centering}|m{1.1cm}<{\centering}m{1.1cm}<{\centering}m{1.5cm}<{\centering}}
\toprule
\raisebox{0pt}[1.5ex][1.5ex]{Method} & \raisebox{0pt}[1.5ex][1.5ex]{AUC} & \raisebox{0pt}[1.5ex][1.5ex]{KS} & \raisebox{0pt}[1.5ex][1.5ex]{Recall@10} \\
\midrule
\raisebox{0pt}[1.5ex][1.5ex]{Pre-service Model} & \raisebox{0pt}[1.5ex][1.5ex]{0.8034} & \raisebox{0pt}[1.5ex][1.5ex]{0.4591} & \raisebox{0pt}[1.5ex][1.5ex]{0.3758} \\
\midrule
\raisebox{0pt}[1.5ex][1.5ex]{MGKD w/o distill} & \raisebox{0pt}[1.5ex][1.5ex]{0.8098} & \raisebox{0pt}[1.5ex][1.5ex]{0.4684} & \raisebox{0pt}[1.5ex][1.5ex]{0.3855} \\
\midrule
\raisebox{0pt}[1.5ex][1.5ex]{\makecell{MGKD w/o \\ fine-grained distill}} & \raisebox{0pt}[1.5ex][1.5ex]{0.8153} & \raisebox{0pt}[1.5ex][1.5ex]{0.4781} & \raisebox{0pt}[1.5ex][1.5ex]{0.3935} \\
\midrule
\raisebox{0pt}[1.5ex][1.5ex]{\makecell{MGKD w/o \\ coarse-grained distill}} & \raisebox{0pt}[1.5ex][1.5ex]{0.8164} & \raisebox{0pt}[1.5ex][1.5ex]{0.4812} & \raisebox{0pt}[1.5ex][1.5ex]{0.3958} \\
\midrule
\raisebox{0pt}[1.5ex][1.5ex]{MGKD} & \raisebox{0pt}[1.5ex][1.5ex]{\textbf{0.8206}} & \raisebox{0pt}[1.5ex][1.5ex]{\textbf{0.4835}} & \raisebox{0pt}[1.5ex][1.5ex]{\textbf{0.3986}} \\
\bottomrule
\end{tabular}
\label{ablation_mgkd}
\end{table}

\subsubsection{The Effects of Pre-training} 
We first examine the impact of pre-training, which in this context refers to leveraging in-service data to enhance the model's initial training, on the model's performance. The \textit{pre-service model} refers to the baseline model trained solely on pre-service data without any pre-training or distillation, while the \textit{MGKD w/o distill} model represents the scenario where in-service data is used for pre-training but without the application of knowledge distillation techniques. As shown in Table~\ref{ablation_mgkd}, the \textit{pre-service model} demonstrates lower AUC, KS, and Recall@10 metrics compared to the models that incorporate pre-training, with drops of 0.80\%, 2.03\%, and 2.58\%, respectively. This indicates that leveraging in-service data during the pre-training phase significantly enhances the model's ability to predict defaults. 

\subsubsection{The Effects of Single Granularity Distillation} 
Next, we evaluate the effects of single granularity distillation by examining two scenarios: removing coarse-grained distillation, namely MGKD w/o coarse-grained distill, and removing fine-grained distillation, namely MGKD w/o fine-grained distill. The results in Table \ref{ablation_mgkd} show that both \textit{w/o coarse-grained distill} and \textit{w/o fine-grained distill} outperform the \textit{pre-service model} and \textit{w/o distill}, indicating that distillation at either level contributes to performance improvement. However, neither approach achieves the same level of performance as our full MGKD framework, suggesting that both coarse-grained and fine-grained distillation play complementary roles in enhancing the model's effectiveness.

When fine-grained distillation is removed, the AUC decreases by 0.64\%, KS by 1.12\%, and Recall@10 by 1.28\% compared to the full MGKD framework. On the other hand, removing coarse-grained distillation results in a smaller decline: AUC decreases by 0.51\%, KS by 0.48\%, and Recall@10 by 0.70\%. This analysis indicates that fine-grained distillation, which aligns the model's output probabilities more closely with the teacher model, is slightly more effective than coarse-grained distillation in improving the model's performance. 

\subsubsection{The Effects of Multi-Granularity Distillation} 
We examined the impact of multi-granularity distillation, specifically fine-grained (output layer) and coarse-grained (hidden layer) distillation. Our model, which incorporates both approaches, achieves the highest AUC, KS, and Recall@10 metrics, as shown in Table \ref{ablation_mgkd}, highlighting that combining distillation at multiple granularities significantly enhances performance by capturing both high-level representations and detailed predictive information from in-service data. Moreover, removing both fine-grained and coarse-grained distillation results in a notable performance drop compared to the full MGKD framework, with AUC deteriorating by 1.30\%, KS by 3.12\%, and Recall@10 by 3.29\%. These declines are larger than when either distillation type is removed individually, emphasizing their complementary roles and the necessity of each in optimizing model performance.

\subsection{Hyperparameter Sensitivity Analysis~(RQ4)}
To answer the fourth research question, we report the sensitivity analysis of the MGKD framework concerning its key hyperparameters, specifically the distillation intensities \(\alpha\) and \(\beta\). These parameters control the extent of knowledge transfer at different levels. We vary each hyperparameter while keeping others fixed to evaluate the impact on the model's performance. The results are presented in Figure~\ref{sensitivity}.

\subsubsection{Distillation Intensity \(\alpha\)}
As shown in all three plots of Figure~\ref{sensitivity}, varying the \(\alpha\) parameter, which controls the output layer distillation, has a significant impact on the model's performance across AUC, KS, and Recall@10. The performance metrics improve as \(\alpha\) increases from 0 to 0.15, reaching their peak at \(\alpha = 0.15\). This suggests that a moderate level of distillation at the output layer may enhance the model's ability to distinguish between classes, leading to improved predictive performance. However, after \(\alpha = 0.15\), all three metrics begin to decline. This indicates that while some level of distillation at the output layer appears beneficial, excessive distillation may lead to overfitting, where the model becomes too tailored to the training data and loses its ability to generalize effectively to new data.

\subsubsection{Distillation Intensity \(\beta\)}
The plots in Figure~\ref{sensitivity} also illustrate the effect of the \(\beta\) parameter, which controls distillation at the hidden layers. Similar to \(\alpha\), the performance metrics (AUC, KS, and Recall@10) increase as \(\beta\) is raised from 0 to 0.20. The peak performance occurs at \(\beta = 0.20\), beyond which the metrics begin to decline. This pattern suggests that a carefully tuned \(\beta\) contributes positively to the model's internal representation, leading to better performance. However, increasing \(\beta\) beyond the optimal point seems to cause a decrease in performance, likely due to over-regularization. This finding highlights the role of \(\beta\) in ensuring the robustness of the MGKD framework, particularly in dynamic online environments as highlighted in RQ2.

\section{Conclusion}\label{sec:conclude}

In this paper, we investigate the integration of pre-service and in-service data for financial risk forecasting and propose a novel framework to enhance the performance of risk assessment before service activation. The proposed model utilizes a knowledge distillation framework where a teacher model trained on in-service data guides a student model trained on pre-service data. 
We introduce a multi-granularity distillation strategy, including coarse-grained, fine-grained, and self-distillation. 
We perform extensive experiments on real-world industrial datasets to evaluate the performance of our model. The experimental results show the superiority of the MGKD approach in both offline and online settings.



\section*{Acknowledgment}

This work was supported by the National Key R\&D Program of China (No. 2023YFC3304701), the National Science Foundation of China (No. 62272466, U24A20233, U2436209), and the Tencent Rhino-Bird Focused Research Program.


\newpage


\newpage

\appendix

\section{Experimental Details}
In this appendix section, we provide a comprehensive overview of the experimental setup for evaluating the MGKD framework. We begin by detailing the metrics used and explaining the rationale behind selecting these metrics to measure performance. Following this, we present the hyperparameter settings and design, highlighting the rigorous hyperparameter search conducted to optimize the MGKD framework's performance.

\subsection{Metrics}
We use three widely adopted metrics to measure the performance of financial default detection, namely AUC, KS, and Recall@10.

The first metric, AUC (Area Under the ROC Curve), is defined as:
\begin{equation}
AUC = \frac{\sum_{u\in U^+}\mathrm{rank}_u - \frac{|U^+|\times(|U^+|+1)}{2}}{|U^+|\times|U^-|}
\end{equation}
Here, $U^+$ and $U^-$ denote the positive and negative sets in the testing set, respectively. Moreover, $\mathrm{rank}_u$ indicates the rank of user $u$ based on the prediction score.

The second metric, KS (Kolmogorov-Smirnov), measures the maximum difference between the cumulative distribution functions of the positive and negative scores. It is calculated as:
\begin{equation}
KS = \max_x \left| F_1(x) - F_0(x) \right|
\end{equation}
where $F_1(x)$ and $F_0(x)$ are the cumulative distribution functions of the positive and negative scores, respectively.

The third metric, Recall@10, indicates the recall when considering the top 10\% of the ranked predictions. Given the low positive rate in mobile payment services (around 0.5\% in our dataset), this metric is particularly important. It reflects the ability to detect top-ranked positive samples and balance the impact on the real-world business system. It is defined as:
\begin{equation}
Recall\text{@}10 = \frac{|\text{TP} \cap \text{Top 10\%}|}{|\text{TP}|}
\end{equation}
Here, $ \text{TP} \cap \text{Top 10\%} $ represents the set of true positives (TP) that are within the top 10\% of the predicted scores. $|\text{TP}|$ denotes the total number of true positives. Recall@10 measures the proportion of true positives that are ranked within the top 10\% of all predictions.

\subsection{Hyperparameter Design}

The MGKD framework was optimized using an extensive hyperparameter search. Hidden channels were selected from \{64, 128, 256, 512\}, and layer counts varied among \{2, 3, 4, 6, 8\}. The learning rate was adjusted within \{0.0001, 0.0005, 0.001, 0.005, 0.01\} using the Adam optimizer. It was complemented by a weight decay of $\{0, 1\times 10^{-7}, 1\times 10^{-6}, 1\times 10^{-5}, 1\times 10^{-4}\}$ and dropout rates from \{0.0, 0.1, 0.2, 0.3, 0.4, 0.5\}. Training was conducted for up to 100 epochs, with early stopping triggered after 50 epochs without improvement.

For knowledge distillation, $\alpha$ was selected from \{0.0, 0.1, 0.2, 0.3, 0.4\}, $\beta$ from \{0.0, 0.05, 0.1, 0.15, 0.2, 0.25, 0.3\}, and the temperature $\tau$ ranged among \{1.0, 1.5, 2.0, 2.5, 3.0\}. These parameters were chosen to maximize AUC, KS, and Recall@10, ensuring model robustness across varying conditions. The optimal hyperparameter settings for the MGKD framework are displayed in Table~\ref{optimal_hyperparameters}.

\begin{table}[!t]
\centering
\caption{Optimal Hyperparameter Settings}
\renewcommand{\arraystretch}{1.5} 
\begin{tabular}{c|c}
\toprule
\textbf{Parameter} & \textbf{Optimal Value} \\
\midrule
hidden channels & 256 \\
num layers & 2 \\
learning rate & 0.005 \\
weight decay & $1\times 10^{-7}$ \\
dropout & 0.4 \\
$\alpha$ & 0.2 \\
$\beta$ & 0.25 \\
$\tau$ & 2.5 \\
\bottomrule
\end{tabular}
\label{optimal_hyperparameters}
\end{table}

\section{Addressing Class Imbalance}

In real-world financial risk prediction scenarios, the distribution of positive (default) and negative (non-default) samples is often highly imbalanced, where the minority class (default) is significantly underrepresented. To address this challenge and mitigate the model's bias towards the majority class, we adopt two complementary strategies: a re-weighting strategy and focal loss. These techniques aim to enhance the model's sensitivity to the minority class while maintaining overall performance.

\subsection{Re-weighting Strategy}

The re-weighting strategy adjusts the loss function to assign higher weights to the minority class samples. Specifically, the weights are computed based on the class distribution:
\begin{equation}
w_c = \frac{1}{\pi_c}
\end{equation}
where $w_c$ represents the weight for class $c \in \{0, 1\}$, and $\pi_c$ denotes the proportion of samples belonging to class $c$ in the training data. The weighted loss for binary classification is then defined as:
\begin{equation}
\mathcal{L}_{\text{weighted}} = -\sum_{i=1}^N w_{y_i} \cdot \Big[ y_i \log \hat{y}_i + (1 - y_i) \log (1 - \hat{y}_i) \Big]
\end{equation}
where $y_i \in \{0, 1\}$ is the true label for sample $i$, $\hat{y}_i$ is the predicted probability for the positive class, and $w_{y_i}$ is the weight for the corresponding class, determined by the class distribution. By applying larger weights to the minority class, the model learns to prioritize minority class samples during training, thereby reducing the impact of class imbalance on predictive performance.

\end{document}